\documentclass[a4paper,twoside]{article}

\usepackage{epsfig}
\usepackage{subcaption}
\usepackage{calc}
\usepackage{amssymb}
\usepackage{amstext}
\usepackage{amsmath}
\usepackage{amsthm}
\usepackage{multicol}
\usepackage{pslatex}
\usepackage{natbib}
\let\bibhang\relax
\usepackage{apalike}
\usepackage{arydshln}
\usepackage{url}
\usepackage{amsmath}

\DeclareMathOperator*{\argmin}{\arg\!\min}

\newlength{\bibitemsep}
\newlength{\bibparskip}
\let\oldthebibliography\thebibliography
\renewcommand\thebibliography[1]{%
  \oldthebibliography{#1}%
  \setlength{\parskip}{\bibitemsep}%
  \setlength{\itemsep}{\bibparskip}%
}
\usepackage{enumitem}
\usepackage{algorithm2e}
\usepackage[bottom]{footmisc}
\usepackage{SCITEPRESS}     

\begin{document}

\title{Adapter-based Approaches to Knowledge-enhanced Language Models - A Survey}

\author{\authorname{Alexander Fichtl\sup{1}, Juraj Vladika\sup{2} and Georg Groh\sup{3}}
\affiliation{\sup{1,3}Social Computing Research Group, Technical University of Munich, Boltzmannstße 3, 85748, Garching, Germany}
\affiliation{\sup{2}SEBIS, Technical University of Munich, Boltzmannstraße 3, 85748, Garching, Germany}
\email{\{alexander.fichtl, juraj.vladika, georg.groh\}@tum.de}}

\keywords{Natural Language Processing, Knowledge Engineering, Knowledge Enhancement, Knowledge Graphs, Language Models, Adapters}

\abstract{Knowledge-enhanced language models (KELMs) have emerged as promising tools to bridge the gap between large-scale language models and domain-specific knowledge. KELMs can achieve higher factual accuracy and mitigate hallucinations by leveraging knowledge graphs (KGs). They are frequently combined with adapter modules to reduce the computational load and risk of catastrophic forgetting. In this paper, we conduct a systematic literature review (SLR) on adapter-based approaches to KELMs. We provide a structured overview of existing methodologies in the field through quantitative and qualitative analysis and explore the strengths and potential shortcomings of individual approaches. We show that general knowledge and domain-specific approaches have been frequently explored along with various adapter architectures and downstream tasks. We particularly focused on the popular biomedical domain, where we provided an insightful performance comparison of existing KELMs. We outline the main trends and propose promising future directions.}

\onecolumn \maketitle \normalsize \setcounter{footnote}{0} \vfill

\section{\uppercase{Introduction}}
\label{sec:introduction}

The field of natural language processing (NLP) has, in recent years, been dominated by the rise of large language models (LLMs). These models are usually pre-trained on large amounts of unstructured textual data, which enables them to solve complex reasoning tasks and generate new text. Still, LLMs can lack awareness of structured knowledge hierarchies, such as relations between concepts and reasoning capabilities in knowledge-intensive tasks \citep{rosset_2020, hu2023}. These drawbacks can lead to inaccurate predictions in downstream tasks and so-called "hallucinations" \citep{huang_2023} within text generation, making LLMs less reliable in practice, an especially precarious issue in high-risk domains like healthcare or law. 

A potential solution to counteract hallucinations and improve the reliability of LLMs is knowledge enhancement: By leveraging expert knowledge from sources such as manually curated knowledge graphs (KGs), structured knowledge can be injected into LLMs. Such knowledge-enhanced language models (KELMs) are a promising approach for higher structured knowledge awareness, better factual accuracy, and less hallucinations \citep{ColonHernandez21, Wei21, hu2023}. KGs are a vital part of \textit{knowledge engineering}, a discipline that can be leveraged to make LLMs use advanced logic and formal expressions \citep{allen_et_al:TGDK.1.1.3}.

Unfortunately, knowledge enhancement in the form of supervised fine-tuning (SFT) can be highly computationally expensive, especially for LLMs with billions of parameters. A promising research avenue to overcome this limitation is using lightweight and efficient adapter modules \citep{houlsbyadapter, pfeifferadapter}. These modules can enhance the task performance of LLMs and, at the same time, be a very computationally efficient solution. Despite the rising popularity of this approach and to the best of our knowledge, a comprehensive overview of adapter-based KELMs is still missing in the NLP research landscape.

To bridge this research gap, we conduct a systematic literature review (SLR) on adapter-based knowledge enhancement of LLMs. Our contributions are (1) a novel review of adapter-based knowledge enhancement, (2) a quantitative and qualitative analysis of different methods in the field, and (3) a detailed categorization of literature and identification of the most promising trends. This work is a reworked and updated version of an SLR conducted as part of a master's thesis of one of the authors \citep{mythesis}. 

\section{Background and Related Work}
\label{sec:background}
This section gives an overview of related work and existing surveys on knowledge enhancement. Knowledge graphs are the most common external knowledge source, so we start with their overview.

\subsection{Knowledge Graphs}

\textbf{Knowledge graphs} (KGs) are a structured representation of world knowledge and have seen a rising prominence in NLP research over the past decade \citep{schneider22}. \citet{Hogan20} define a KG as "\textit{a graph of data intended to accumulate and convey knowledge of the real world, whose nodes represent entities of interest and whose edges represent relations between these entities}". Similarly, \citet{ji21} published a comprehensive survey on KGs and, following existing literature, defined the concept of a KG as "$\mathcal{G}=\{\mathcal{E}, \mathcal{R}, \mathcal{F}\}$, \textit{where} $\mathcal{E}, \mathcal{R}$ \textit{and} $\mathcal{F}$ \textit{are sets of entities, relations and facts, respectively; a fact is denoted as a triple} $(h, r, t) \in \mathcal{F}$". $h, r, t$ denote the head, relation, and tail of a triplet. An \textbf{Ontology} is a formal representation of concepts and semantic relations between them -- it provides a "schema" that a KG has to adhere to, making it possible to do reasoning and derive rules from KGs \citep{KHADIR2021100339}.

Depending on the source and purpose of a KG, entities and relations can take on various shapes. For example, in the biomedical knowledge graph UMLS \citep{umls}, a relation can take the shape of a single word like "inhibits", a short phrase like "relates to", or a compound term including, for example, chemical or medical categories such as "[protein] relates to [disease]" or "[substance] induces [physiology]". A textual connection is vital because it links the graph structure with natural language, simplifying the integration of information from KGs into language models and the associated learning processes. Other than UMLS, other examples of popular KGs are DBpedia \citep{dbpedia} and ConceptNet \citep{speer17}.

\subsection{Approaches to Knowledge Enhancement}

\begin{figure*}[htb!]
  \centering
  \includegraphics[width=0.77\textwidth]{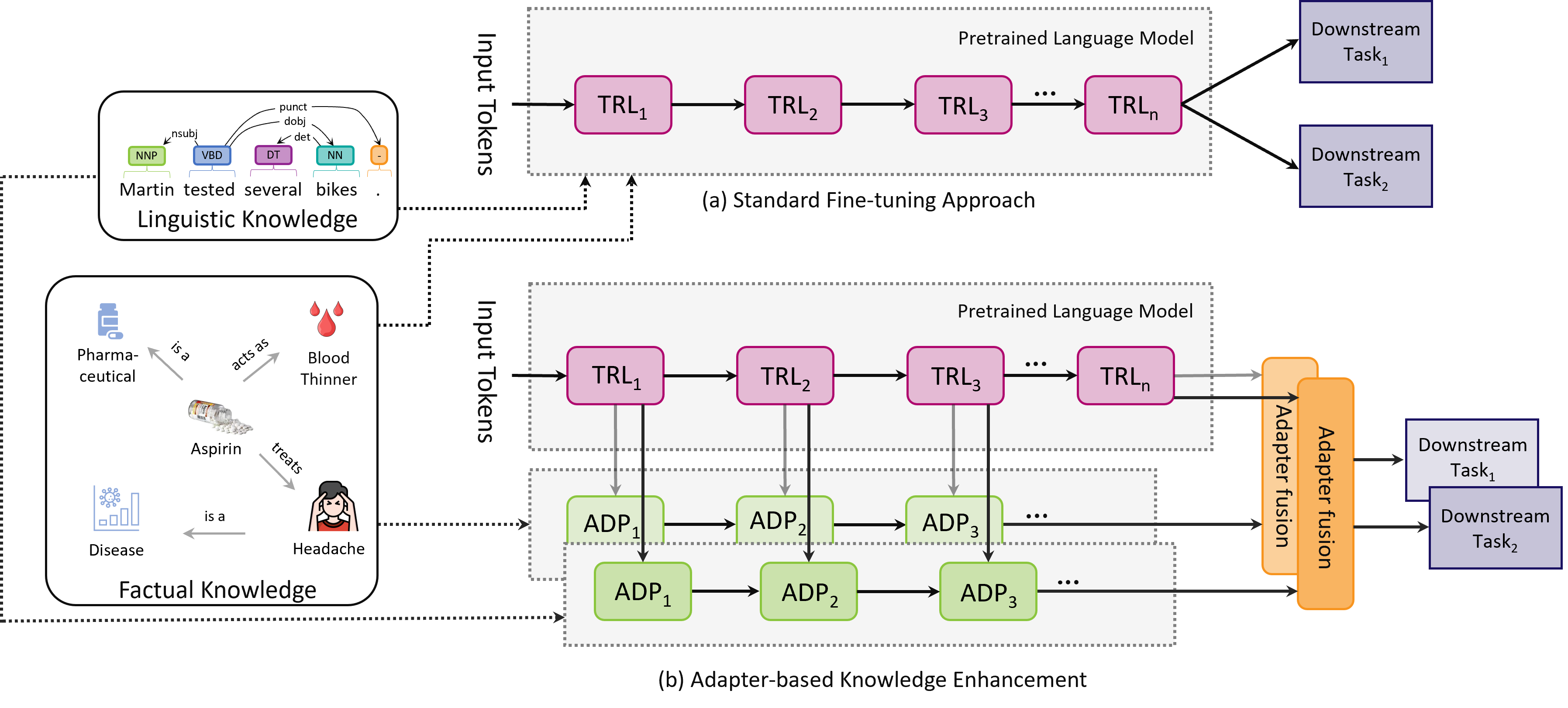}
  \caption{Illustration of a standard fine-tuning versus a knowledge enhancement process. In the example, knowledge from a KG is injected into the model via adapters.} \label{fig:knowledge_enhancement}
\end{figure*}

At the time of writing, some reviews had already been published that gave an overview of KELMs and classified different approaches. \citet{ColonHernandez21} review the existing literature and split the approaches to integrate structure knowledge with LMs into three categories: (1) input-centered strategies, centering around altering the structure of the input or selected data, which is fed into the base LLM; (2) architecture-focused approaches, which involve either adding additional layers that integrate knowledge with the contextual representations or modifying existing layers to alter parts like attention mechanisms; (3) output-focused approaches, which work by changing either the output structure or the losses used in the base model. Our study focuses on category (2) by examining the adapter-based mechanisms for injecting information into the model (see example in Figure \ref{fig:knowledge_enhancement}), which were shown to be the most promising by the authors.


The second survey by \citet{Wei21} reviews a large number of studies on KELMs and classifies them using three taxonomies: (1) knowledge sources, (2) knowledge granularity, and (3) application areas. Within (1), the knowledge sources include linguistic, encyclopedic, common-sense, and domain-specific knowledge. 
The second taxonomy (2) acknowledges the common approach of using KGs as a source of knowledge. 
Levels of granularity mentioned are text-based knowledge, entity knowledge, relation triples, and KG sub-graphs. 
Lastly, with the third taxonomy (3), the authors discuss how knowledge enhancement can improve natural language generation and understanding. They also review popular benchmarks that can be used for task evaluation of KELMs \citep{Wei21}.

Adapters are part of a broader paradigm of \textit{modular deep learning}, described in detail by \citet{pfeiffer2024modulardeeplearning}. The two field studies by \citet{ColonHernandez21} and \citet{Wei21} on the classification of KELM approaches were a valuable starting point for exploring KELMs. Although they address some adapter-based studies like K-Adapter \citep{kadapter}, most other adapter-based KELMs are missing.
This lack of coverage led us to conduct a novel systematic literature search focusing specifically on the adapter-based KELMs, considering their rising popularity and importance.

\section{Adapters}

This section will give an overview of LLM adapters and their applications to establish a conceptual understanding of adapter-based enhancement.

\subsection{Overview}

Broadly speaking, adapters are small bottleneck feed-forward layers inserted within each layer of an LLM \citep{houlsbyadapter}. The small number of additional parameters allows for injecting new data or knowledge without fine-tuning the whole model. This feat is usually accomplished by freezing the layers of the large base model  while only updating the adapter weights. Due to the lightweight nature of adapters, this approach leads to short training times with relatively low computing resource requirements. Adapters were mainly used for quick and cheap downstream-task fine-tuning but are now increasingly used for knowledge enhancement. Because it is possible to train adapters individually, they can also be used for multi-task training by specializing one adapter for each task or multi-domain knowledge injection by specializing adapters to different domains \citep{pfeifferadapter}. 

Leveraging adapters in LLMs also has positive "side effects": Adapters can avoid catastrophic forgetting (the issue when an LLM suddenly deteriorates in performance after fine-tuning) by introducing new task-specific parameters \citep{houlsbyadapter, pfeifferadapter} and, in transfer learning, adapters have even been shown to improve stability and adversarial robustness for downstream tasks \citep{han2021}. Additionally, adapters have been shown to operate on top of the base model’s frozen representation space while largely preserving its structure rather than on an isolated subspace \citep{alabi-etal-2024-hidden}.

\subsection{Adapter Types}

\paragraph{Houlsby Adapter}

\begin{figure}[ht!]
  \centering
  \includegraphics[width=0.47\textwidth]{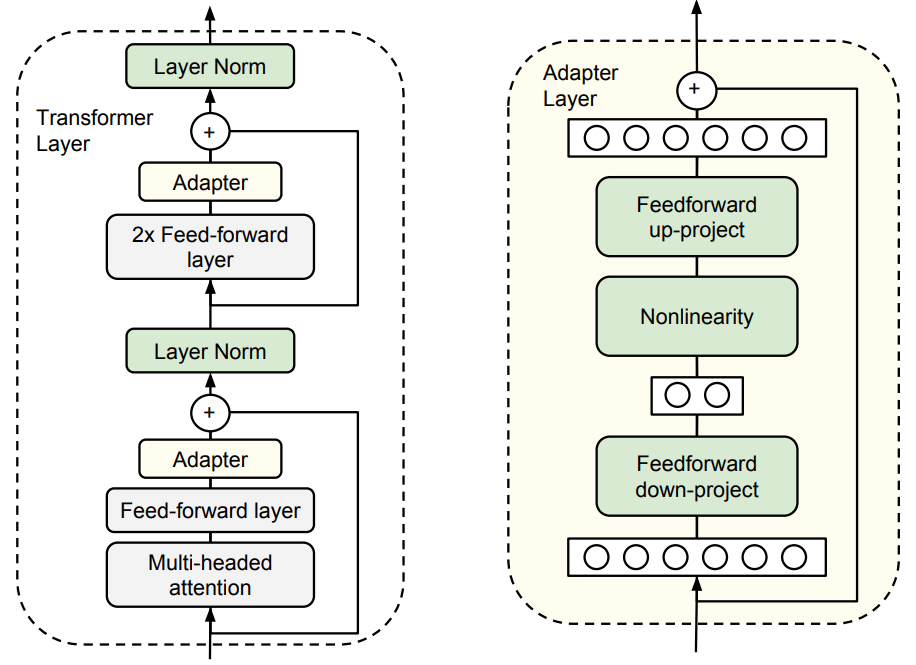}
  \caption{Location of the adapter module in a transformer layer (left) and architecture of the Houlsby Adapter (right). All green layers are trained on fine-tuning data, including the adapter itself, the layer normalization parameters, and the final classification layer (not shown). Image with permission from \citet{houlsbyadapter}.} \label{fig:houlsby} 
\end{figure}

The Houlsby Adapter \citep{houlsbyadapter} was the first adapter to be used for transfer learning in NLP. The idea was based on adapter modules initially introduced by \citet{rebuffi17} in the computer vision domain. The two main principles stayed the same: Adapters require a relatively small number of parameters compared to the base model and a near-identity initialization. These principles ensure that the total model size grows relatively slowly when more transfer tasks are added, while a near-identity initialization is required for stable training of the adapted model \citep{houlsbyadapter}. The optimal architecture of the Houlsby Adapter was determined by meticulous experimenting and tuning; the result can be seen in figure \ref{fig:houlsby}. In a classical transformer structure \citep{vaswani2017}, the adapter module is added once after the multi-headed attention and once after the two feed-forward layers. The modules project the $d$-dimensional layer features of the base model into a smaller dimension, $m$, then apply a non-linearity (like ReLU) and project back to $d$ dimensions. The configuration also hosts a skip-connection, and the output of each sub-layer is forwarded to a layer normalization \citep{layernorm}. Including biases, $2md + d + m$ parameters are added per layer, accounting for only 0.5 to 8 percent of the parameters of the original BERT model used by the authors when setting $m << d$.

\paragraph{Bapna and Firat Adapter}

In contrast to the Houlsby Adapter, \citet{bapnafiratadapter} only introduce one adapter module in each transformer layer: they keep the adapters after the multi-headed attention (so-called "top" adapters) while dropping the adapters after the feed-forward layers (so-called "bottom" adapters) of the transformer (refer to Figure \ref{fig:houlsby} for better understanding of the component positions). Moreover, while \citet{houlsbyadapter} re-train layer normalization parameters for every domain, \citet{bapnafiratadapter} "simplify this formulation by leaving the parameters frozen, and introducing new layer normalization parameters for every task, essentially mimicking the structure of the transformer feed-forward layer".

\paragraph{Pfeiffer Adapter and AdapterFusion.}

The approaches of \citet{bapnafiratadapter, houlsbyadapter} did not allow information sharing between tasks. \citet{pfeifferadapter} introduce Adapter Fusion, a two-stage algorithm that addresses the sharing of information encapsulated in adapters trained on different tasks. In the first stage, they train the adapters in single-task or multi-task setups for $N$ tasks similar to the Houlsby Adapter, but only keeping the top adapters, similar to the Bapna and Firat Adapter. As a second step, they combine the set of $N$ adapters with AdapterFusion: They fix the parameters $\Theta$ and all adapters $\Phi$, and finally introduce parameters $\Psi$ that learn to combine the $N$ task adapters for the given target task \citep{pfeifferadapter}: $
\Psi_m \leftarrow \underset{\Psi}{\argmin}\:L_m\left(D_m ; \Theta, \Phi_1, \ldots, \Phi_N, \Psi\right)
$

Here, $\Psi_m$ are the learned AdapterFusion parameters for task $m$. In the process, the training dataset of $m$ is used twice: once for training the adapters $\Phi_m$ and again for training Fusion parameters $\Psi_m$, which learn to compose the information stored in the $N$ task adapters \citep{pfeifferadapter}. With their approach of separating knowledge extraction and knowledge composition, they further improve the ability of adapters to avoid catastrophic forgetting and interference between tasks and training instabilities.
The authors also find that using only a single adapter after the feed-forward layer performs on par with the Houlsby adapter while requiring only half of the newly introduced adapters \citep{pfeifferadapter}. This fact makes the Pfeiffer adapter an attractive choice for many applications, further proven by its popularity among the papers in our review.

\paragraph{K-Adapter}
\citet{kadapter} follow a substantially different approach where the adapters work as "outside plug-ins". In their work, an adapter model consists of $K$ adapter layers (hence the name) that contain $N$ transformer layers and two projection layers. Similar to the approaches above, a skip connection is added but instead applied across the two projection layers. The adapter layers are plugged in among varying transformer layers of the pre-trained model. The authors explain that they concatenate the output hidden feature of the transformer layer in the pre-trained model and the output feature of the former adapter layer as the input feature of the current adapter layer. 

\paragraph{}
Adapter architectures for knowledge enhancement exist that differ from the four adapter types mentioned here. For example, the "Parallel Adapter" \citep{paralleladapter} or the adapter architecture by \citet{stickland19}). However, as the upcoming comprehensive literature survey will show, these architectures are either unique to specific papers or have not found broader applications in the field of KELMs. Another popular type of efficient adaptation we'd like to mention for completeness is low-rank adaptation or LoRA \citep{hu2022lora} and its quantized version QLoRA \citep{dettmers2023qlora}. These approaches do not add new layers but rather enforce a low-rank constraint on the weight updates of the base model's layers. This methodology enables efficient fine-tuning of LLMs and also allows for domain adaption or knowledge enhancement with KGs \citep{tian-etal-2024-kg}.

\section{Methodology}
\label{section:methods}

This chapter details the methodology we employed for the systematic literature review. We largely followed the procedure of \citet{kitchenham09} for systematic literature reviews in software engineering. The search strategy for the systematic literature review of this study included literature that fulfilled the following inclusion criteria:

\begin{itemize}
\setlength\itemsep{0em}
    \item Peer-reviewed articles from ACM\footnote{\url{https://dl.acm.org/}}, ACL\footnote{\url{https://aclanthology.org/}}, and IEEE Xplore\footnote{\url{https://ieeexplore.ieee.org/Xplore/home.jsp}}
    \item Article abstracts that match the search string \textit{("adapter" OR "adapter-based") AND ("language model" OR "nlp" OR "natural language processing") AND ("injection" OR "knowledge")}
    \item Articles published after February 2, 2019 (publication of the Houlsby Adapter, the first LLM adapter)
    \item Articles that address the topic of adapter-based knowledge-enhanced language models
\end{itemize}

We also included three articles not found in the databases because they were fundamental works on the topic of the SLR and frequently referenced. The SLR was concluded in January 2024 and represents the state of research literature up to this point. Additional details on the methodology can be found in Appendix \ref{sec:appendix_methodology}


\section{Results}
\label{section:results}

This section will present the results of the SLR on adapter-based knowledge enhancement.

\subsection{Overview}

\begin{table}[!htb]
\footnotesize
    \centering
    \footnotesize
    \begin{tabular}{llll}
    \hline
        \textbf{Source} & \textbf{Initial} & \textbf{Abstract} & \textbf{Full Text} \\ \hline
        \textbf{IEEE} & 28 & 6 & 6 \\ 
        \textbf{ACM} & 10 & 6 & 5 \\ 
        \textbf{ACL} & 36 & 16 & 13 \\ 
        \textbf{Others} & 2 & 2 & 2 \\ \hdashline
        \textbf{Total} & 76 & 30 & 26 \\
    \end{tabular}
    \caption{\label{tab:survey_results_table} Quantitative overview of the literature sources and the selection process}
\end{table}

\begin{table*}[!htb]
    \centering
    \small
    \footnotesize
    \begin{tabular}{p{56mm}llllll}
    \hline
        \textbf{paper \& nickname} & \textbf{adapter type} & \textbf{scope} & \textbf{main source} & \textbf{task} \\ \hline
        \textbf{K-MBAN \citep{zou22}} & K-Adapter & open & T-REx \scriptsize (Wiki) & RC \\ 
        \textbf{/ \citep{moon21}} & Houlsby & open & WMT20 & MT \\ 
        \textbf{CSBERT \citep{yu23}} & Unique & open & diverse & SL \\ 
        \textbf{/ \citep{qian22}} & Unique & open & AESRC2020 & SR \\
        \textbf{/ \citep{li23}} & Houlsby & closed (multiple) & diverse & SF \\ 
        \textbf{CPK \citep{liu23}} & K-Adapter & closed (biomed) & Wikipedia & RC, ET, QA \\ \hdashline
        \textbf{CKGA \citep{lu23}} & Unique & open & DBpedia & SC \\ 
        \textbf{/ \citep{nguyen23}} & Pfeiffer & open & diverse & SA \\ 
        \textbf{KEBLM \citep{keblm}} & Pfeiffer & closed (biomed) & UMLS & QA, NLI, EL \\ 
        \textbf{/ \citep{guo22}} & Unique & open & Ch. Lexicon & NER \\
        \textbf{/ \citep{tiwari23}} & Unique & closed (biomed) & Vis-MDD &  TS \\ \hdashline
        \textbf{AdapterSoup \citep{chronopoulou23}} & Bapna and Firat & closed (multiple) & diverse &  LM \\ 
        \textbf{/ \citep{wold22}} & Houlsby & open & ConceptNet & LAMA \\ 
        \textbf{/ \citep{chronopoulou22}} & Unique & closed  (multiple) & diverse & LM \\ 
        \textbf{DS-TOD \citep{hung22}} & Pfeiffer & closed (multiple) & CCNet & TOD \\ 
        \textbf{/ \citep{emelin22}} & Houlsby & closed (multiple) & MultiWOZ & TOD \\ 
        \textbf{KnowExpert \citep{xu22}} & Bapna and Firat & open & WoW & KGD \\ 
        \textbf{mDAPT \citep{kaer21}} & Pfeiffer & closed (multiple) & WMT20 & NER, STC \\ 
        \textbf{DAKI \citep{lu21}} & K-Adapter & closed (biomed) & UMLS & NLI \\ 
        \textbf{/ \citep{majewska21}} & Pfeiffer & open & VerbNet & EE \\ 
        \textbf{/ \citep{lauscher20}} & Houlsby & open & ConceptNet & GLUE \\
        \textbf{TADA \citep{hung23}} & Unique & open & CCNet & TOD, NER, NLI \\ 
        \textbf{LeakDistill \citep{vasylenko23}} & StructAdapt & open & AMR graph & SMATCH \\ 
        \textbf{MixDA \citep{diao23}} & Houlsby, Pfeiffer & closed (multiple) & diverse & GLUE, TXM \\ \hdashline
        \textbf{MoP \citep{mop}} & Pfeiffer & closed (biomed) & UMLS & BLURB \\ 
        \textbf{K-Adapter \citep{kadapter}} & K-Adapter & open & T-REx \scriptsize (Wiki) & RCL, ET, QA \\ \hline
    \end{tabular}
    \caption{\label{tab:survey_results} Overview of the results for the literature survey, including all papers and their references. The task and source acronyms are explained in the appendix. The dotted lines separate the database sources: First come the IEEE papers, then ACM, ACL, and finally, the papers from other sources. For the definition of all task acronyms, see Appendix \ref{sec:appendix_acronyms}}
\end{table*}

Table \ref{tab:survey_results_table} shows the source distribution for all included papers. Fifty-nine papers were found by applying the search string as a query on the ACL, ACM, and IEEE search engines. Due to their importance for the field, we included three additional papers from other sources. These papers were found through online search and paper references during the general research process. In summary, after the abstract screening, 31 articles met all inclusion criteria (and no exclusion criteria). After the full paper screening, 26 papers formed the final paper pool of the survey. 
Table \ref{tab:survey_results} gives an overview of all papers included in the survey. It includes information on the adapter type used in the paper, the domain and scope of the paper, and the downstream NLP tasks for which it was developed.

\subsection{Data Analysis}

We will now give a quantitative analysis showcasing and interpreting quantitative distributions, followed by significant qualitative insights from the papers.

\subsubsection{Quantitative Analysis}

\begin{figure}[htb!]
  \centering
  \includegraphics[width=0.4\textwidth]{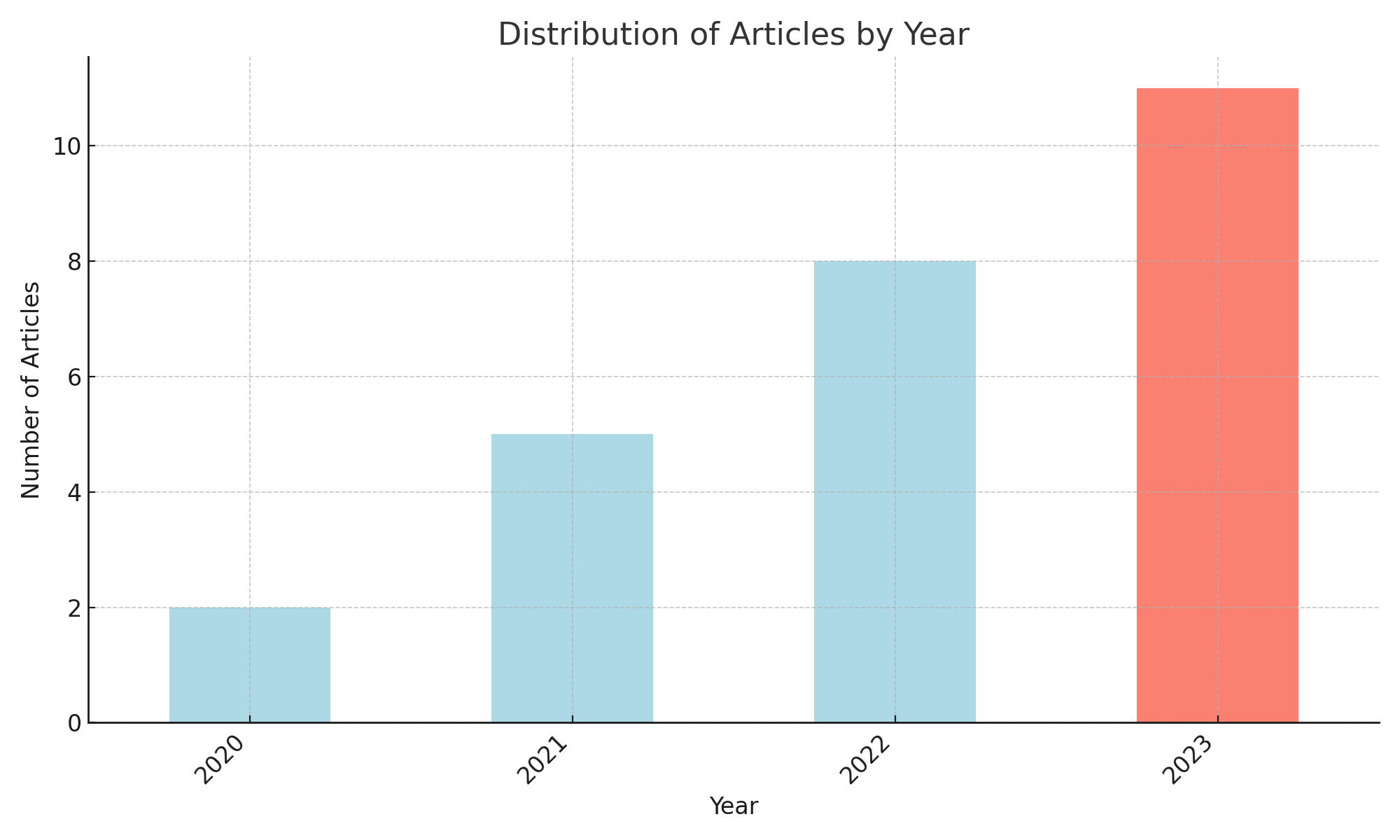}
  \caption{Yearly distribution of publications} \label{fig:year}
\end{figure}

\paragraph{Yearly Distribution}
There has been a significant increase in publications on adapter-based approaches to knowledge-enhanced language models in recent years (Fig. \ref{fig:year}). While only two papers were published in 2020, eleven new papers were published in 2023. This trend suggests growing interest and research activity in the domain.

\begin{figure}[htb!]
  \centering
  \includegraphics[width=0.42\textwidth]{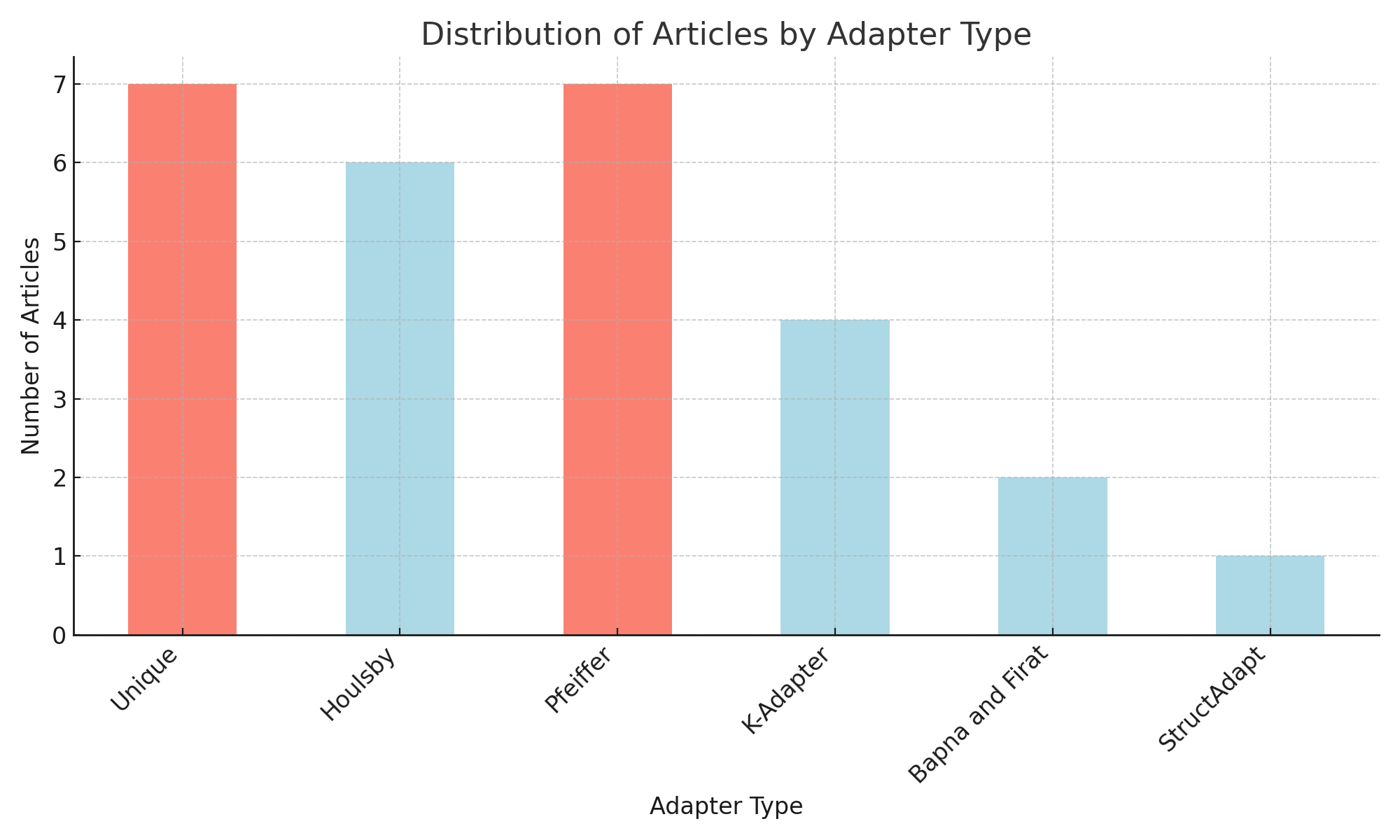}
  \caption{Distribution of adapter types used in the papers} \label{fig:adapters}
\end{figure}

\paragraph{Adapter Type Distribution}
Next, we evaluate the popularity and variety of adapter types used across the papers (Fig. \ref{fig:adapters}). The “Pfeiffer” and "Houlsby" adapter types stand out as the most common, which suggests that the closely related underlying architecture is the most popular methodology in the field. This popularity is likely not only an achievement of the adapter's performance but also due to the well-established Adapter-Hub platform \citep{adapterhub}, which, although offering other options, uses adapters with the Pfeiffer configuration by default. This finding showcases a need and trend to build custom adapters well-suited to individual tasks. In the upcoming years, we will likely see many novel adapter architectures. The “K-Adapter” and “Bapna and Firat” adapters are the less frequently mentioned architectures, suggesting that these approaches are less well-established.

\paragraph{Domain Analysis}
Third, we analyze the distribution of papers across the domain scope and coverage to understand domain-specific preferences in the literature. The open-domain scope is the most popular, with many papers exploring adapter-based approaches within the open domain. The popularity is likely caused by the interest in creating LLMs with a common-secommon-sensending or world knowledge. Furthermore, the single- and multi-domain approaches are split evenly within the closed-domain papers. 
Finally, only six papers focus on the biomedical domain, but relative to other domains, the biomedical field is by far the most prominent of all domain-specific approaches. The popularity likely comes down to the availability of large biomedical KGs, and medicine historically being one of the most active research fields in general science \citep{cimini14}.

\paragraph{Task and Source Distribution}
A highly diverse range of tasks and sources is being explored throughout the papers, which signifies the versatility and potential of adapter-based approaches across different NLP tasks and domains. Tasks such as Reading Comprehension (RC), Named Entity Recognition (NER), and Question Answering (QA) appear to be popular areas of focus in the literature. This could be because these tasks are the most demanding regarding structural knowledge requirements. The knowledge source distributions show only minimal overlap.


\subsubsection{Qualitative Analysis}

\begin{table*}[htpb]
\small
\centering
\begin{tabular}{llllll}
\hline
\textbf{$\downarrow$ model|dataset $\rightarrow$} & \textbf{HoC} & \textbf{PubMedQA} & \textbf{BioASQ7b} & \textbf{MedNLI} & \textbf{NCBI}  \\ \hline
\textbf{SciBERT-base}   & 80.52$_{\pm0.60}$                  & 57.38$_{\pm4.22}$                           & 75.93$_{\pm4.20}$                     & 81.19$_{\pm0.54}$ & 88.57 \\
\textbf{\qquad+ \textit{MoP}}    & 81.79$^\dagger_{\pm0.66}\uparrow$  & 54.66$_{\pm3.10}$                           & 78.50$^\dagger_{\pm4.06}\uparrow$     & 81.20$_{\pm0.37}\uparrow$ & /    \\
\textbf{\qquad+ \textit{KEBLM}}  & /                                           & 59.00$\uparrow$                                       & /                                              & 82.14$\uparrow$ & \textbf{93.50}$\uparrow$     \\
\textbf{\qquad+ \textit{DAKI}}  & /                                           & /                                       & /                                              & / & /    \\
\textbf{\qquad+ \textit{CPK}}  & /                                           & /                                   & /                                              & / & /    \\ \hdashline
\textbf{BioBERT-base}   & 81.41$_{\pm0.59}$                  & 60.24$_{\pm2.32}$                           & 77.50$_{\pm2.92}$                     & 82.42$_{\pm0.59}$ & 88.30   \\
\textbf{\qquad+ \textit{MoP} }   & 82.53$^\dagger_{\pm1.08}\uparrow$  & 61.04$_{\pm4.81}\uparrow$                   & 80.79$^\dagger_{\pm4.40}\uparrow$     & 82.93$_{\pm0.55}\uparrow$ & /    \\
\textbf{\qquad+ \textit{KEBLM}}  & /                                           & \textbf{68.00} $\uparrow$             & /                                              & 84.24 $\uparrow$ & 93.20$\uparrow$ \\
\textbf{\qquad+ \textit{DAKI}}   & /                                           & /            & /                                              & 83.41 $\uparrow$ & 89.01$\uparrow$     \\
\textbf{\qquad+ \textit{CPK} }   & /  & /                 & /    & 81.65 & 88.42$\uparrow$    \\ \hdashline
\textbf{PubMedBERT-base}         & 82.25$_{\pm0.46}$                  & 55.84$_{\pm1.78}$                           & 87.71$_{\pm4.25}$                     & 84.18$_{\pm0.19}$ & 87.82    \\
\textbf{\qquad+ MoP}          & \textbf{83.26}$^\dagger_{\pm0.32}\uparrow$  & 62.84$^\dagger_{\pm2.71}\uparrow$  & \textbf{90.64}$^\dagger_{\pm2.43}\uparrow$ & \textbf{84.70}$_{\pm0.19}\uparrow$ & /    \\ 

\end{tabular}
\caption{performance reports for tasks with highest overlap in the biomedical domain. The metric for HoC is Micro F1; for NCBI, it is F1, while for the other three, it is accuracy. The best results for every task are in bold. "$\uparrow$" denotes that improvements are observed compared to the base model. “$\dagger$" denotes a statistically significant better result over the base model (T-test, p $<$ 0.05), but not all papers report their scores. The baseline performance of the models is taken from the original papers if given. Otherwise, the scores are taken from the MoP results.
} \label{tab:final_results}
\end{table*} 

This section of the analysis highlights recurring themes and individual insights from the papers. Fully summarizing all articles was outside the scope of this survey. However, we still provide an overview of the most common patterns. 

\paragraph{General Knowledge} 

The quantitative analysis showed that open-domain approaches are more popular than their close-domain counterparts. Subsequently, there is also a large variety in the used frameworks, knowledge sources, and overall goals. Two commonly used KGs for general knowledge are ConceptNet \citep{speer17} for common-sense and DBpedia \citep{dbpedia} for encyclopedic world knowledge. Two example works that use these KGs are \citet{wold22} and the CKGA ("knowledge graph-based adapter") by \citet{lu23}. \citet{wold22} train adapter modules on sub-graphs of ConceptNet to inject factual knowledge into LLMs. They evaluate their framework on the Concept-Net Split of the LAMA Probe \citep{petroni19} and see increasing performance while only adding 2.1\% of new parameters to the original models. 
CKGA \citep{lu23} tackles aspect-level sentiment classification by leveraging knowledge from DBpedia. They link aspects to DBpedia end extract an aspect-related sub-graph. Then, a PLM and the KG embedding are utilized to encode the common-sense of entities, where the corresponding knowledge is extracted with graph convolutional networks \citep{lu23}.

\paragraph{Linguistic Knowledge}

Instead of only including factual knowledge, some works inject additional linguistic knowledge into adapters \citep{majewska21, zou22, yu23, kadapter}. While LLMs already encode a range of syntactic and semantic properties of language, \citet{majewska21} explain that LLMs "are still prone to fall back on superficial cues and simple heuristics to solve downstream tasks, rather than leverage deeper linguistic information". Their paper explores the interplay between verb meaning and argument structure. They use knowledge to enhance LLMs with Pfeiffer Adapters to improve English event extraction and machine translation in other languages. Another example is the work of \citet{zou22} on machine reading comprehension (MRC). They proposed the K-MBAN model to integrate linguistic and factual external knowledge into LLMs through K-Adapters.

\paragraph{Domain-specific Knowledge}

\citet{chronopoulou22} propose a parameter-efficient approach to domain adaptation using adapters. They represent domains as a hierarchical tree structure where each node in the tree is associated with a set of adapter weights. Their work focused on specializing adapters in website domains like \textit{booking.com} and \textit{yelp.com}. In another instance, \citet{chronopoulou23} propose "AdapterSoup". In this framework, they also use adapters for domain-specific tasks but use "an approach that performs weight-space averaging of adapters trained on different domains". AdapterSoup can be helpful in various domain-specific approaches in low-resource settings, especially when only a small amount of data on a specific subdomain is obtainable and closely related adapters are available instead. Earlier, we saw that the biomedical domain is the most prevalent among the closed-domain approaches to adapter-based KELMs. We will briefly examine the relevant works in the following.

\paragraph{Biomedical Knowledge}
We have found the works of DAKI \citep{lu21}, MoP \citep{mop}, and KEBLM \citep{keblm} to be the most impactful. According to the results of our literature survey, DAKI ("Diverse Adapters for Knowledge Integration") was the first work to use adapters specifically for knowledge enhancement in the biomedical domain. \citet{lu21} leverage data from the UMLS meta-thesaurus and UMLS Semantic Network groups concepts, but also from Wikipedia articles for diseases as proposed by \citet{he20}. \citet{mop} recognize that KGs like UMLS, which can be several gigabytes large, are very expensive to train on in their entirety. They propose to use a "Mixture of Partitions" (MoP), which splits the KG into sub-graphs and combines later with AdapterFusion \citep{pfeifferadapter}. Finally, the KEBLM framework's trademark is that it allows the inclusion of a variety of knowledge types from multiple sources into biomedical LLMs. In contrast to DAKI, which uses more than one source, KEBLM includes a knowledge consolidation phase after the knowledge injection, where they teach the fusion layers to effectively combine knowledge from both the original PLM and newly acquired external knowledge by using an extensive collection of unannotated texts \citep{keblm}. For completeness, we refer to \citet{kaer21} for information on the m-DAPT framework, which addresses multi-lingual domain adaptation for biomedical LLMs and KeBioSum \citep{kebiosum}, who state their work is the first study exploring knowledge injection for biomedical extractive summarization.

\paragraph{Performance Insights} In the papers covered by this survey, the performance of adapter-based KELMs on downstream tasks is consistently shown to be better than that of base LMs. For example, \citet{diao23} show an increase of +9\% on Common-seCommon-senset{talmor-etal-2019-commonsenseqa} with their mixture-of-adapters approach, while \citet{kaer21} improve financial text classification on OMP-9 \citep{omp} by +4\%. While the task variation across domains is too diverse to be shown systematically in our survey, we report in detail on performance comparison in the biomedical domain in Table \ref{tab:final_results} in the next paragraph.
Another interesting insight is found by \citet{he21}, who show that adapter-based tuning mitigates forgetting issues better than regular fine-tuning since it yields representations with less deviation from those generated by the initial pre-trained language model. 

\paragraph{Performance Comparison (Biomedical)}
\label{sec:performance_biomed}

Table \ref{tab:final_results} gives an overview of the downstream task performance of several papers that are included in this survey. The focus lies on the biomedical domain, so the task overlap is high enough for an insightful comparison. The scores are reported for five downstream tasks, namely HoC \citep{hoc}, PubMedQA \citep{blurb}, BioASQ7b \citep{bioasq7b}, MedNLI \citep{mednli}, and NCBI \citep{NCBI}, as well as three common biomedical language models (SciBERT \citep{scibert}, BioBERT \citep{biobert}, and PubMedBERT \citep{blurb}). While adapter-based KELMs consistently improve performance in almost all instances, performance boosts across different tasks and models vary strongly. In this specific setting, we recommend the MoP \citep{mop} and KEBLM \citep{keblm} frameworks since they show the highest performance boosts (e.g., PubMedQA \citep{pubmedqa} accuracy increase of around +7\% and +8\%, respectively) and overshadow the lower performing CPK and DAKI frameworks in all instances. MoP, in particular, is being continually used for biomedical knowledge enhancement, even in 2024 \citep{icaart24}.

\section{Current and Future Trends}
In this section, we outline the most important findings and trends and point out promising future directions:

\begin{itemize}
\setlength\itemsep{0em}
    \item Adapter-based KELMs are a recent development in NLP, but interest in them is rising fast, with a linear yearly increase of published papers. We predict the growth trend to continue.
    \item Various adapter architectures exist and are advanced by researchers to be more efficient while preserving task performance. This advancement has temporarily peaked with the Pfeiffer adapter, the most popular type. We expect future work to focus their updates on adapter architecture by overcoming the latency of sequential data processing in adapters and enabling hardware parallelism.
    \item Research focuses on the open domain -- injecting general world knowledge into models. Within the closed domain, the biomedical domain is the most popular, owing to the existence of large biomedical KGs. We foresee the potential to apply adapter-based KELMs to other highly structured domains, such as the legal or financial domain (documents with rigid structure).
    \item The largest improvements in task performance is seen in knowledge-intensive tasks like question answering and text classification, with more minor improvements for reasoning tasks like entailment recognition. Generative tasks, other than dialogue modeling, are rather unexplored. We envision a future popular use case that could use knowledge enhancement to improve the factuality and informativeness of generated text.

\end{itemize}

\section{Conclusion}
\label{section:conclusion}

In this paper, we conducted a systematic literature review on approaches to enhancing language models with external knowledge using adapter modules. We portrayed which adapter-based approaches exist and how they compare to each other. We showed there is a steady growth of interest in this domain with each new year and highlighted the most popular adapter architectures (with "Pfeiffer" as the predominant one). We discovered a balance in popularity between open-domain approaches focusing on integrating general world knowledge into models and closed-domain approaches focusing on specialized fields, with biomedical as the most popular domain. With our review, we contribute a novel and extensive resource for this nascent yet fast-growing field, and we hope it will be a useful entry point for other researchers in the future.

\section*{Limitations}

Our literature search methodology follows a strict search string and exclusion criteria. Subsequently, we might have overlooked some relevant work on adapter-based KELMs. Also, some reviewed papers were not adequately analyzed in this work due to space constraints, leading to potentially missing insights and a non-complete representation of the state of research on adapter-based enhancement. Additionally, due to the variety of applications and domains, we could not give precise guidelines on what methods to use under which circumstances. Still, we aimed to report on the most common patterns and trends discovered in the literature, which can serve as a basis for future research.


\footnotesize{\bibliography{example}}

\begin{thebibliography}{}

\bibitem[Alabi et~al., 2024]{alabi-etal-2024-hidden}
Alabi, J., Mosbach, M., Eyal, M., Klakow, D., and Geva, M. (2024).
\newblock The hidden space of transformer language adapters.
\newblock In {\em Proceedings of the 62nd Annual Meeting of the Association for Computational Linguistics}, pages 6588--6607. Association for Computational Linguistics.

\bibitem[Allen et~al., 2023]{allen_et_al:TGDK.1.1.3}
Allen, B.~P., Stork, L., and Groth, P. (2023).
\newblock {Knowledge Engineering Using Large Language Models}.
\newblock {\em Transactions on Graph Data and Knowledge}, 1(1):3:1--3:19.

\bibitem[Auer et~al., 2007]{dbpedia}
Auer, S., Bizer, C., Kobilarov, G., Lehmann, J., Cyganiak, R., and Ives, Z. (2007).
\newblock Dbpedia: A nucleus for a web of open data.
\newblock In {\em The Semantic Web}, pages 722--735, Berlin, Heidelberg. Springer Berlin Heidelberg.

\bibitem[Ba et~al., 2016]{layernorm}
Ba, J., Kiros, J.~R., and Hinton, G.~E. (2016).
\newblock Layer normalization.
\newblock {\em ArXiv}, abs/1607.06450.

\bibitem[Baker et~al., 2015]{hoc}
Baker, S., Silins, I., Guo, Y., Ali, I., Högberg, J., Stenius, U., and Korhonen, A. (2015).
\newblock {Automatic semantic classification of scientific literature according to the hallmarks of cancer}.
\newblock {\em Bioinformatics}, 32(3):432--440.

\bibitem[Bapna and Firat, 2019]{bapnafiratadapter}
Bapna, A. and Firat, O. (2019).
\newblock Simple, scalable adaptation for neural machine translation.
\newblock In {\em Proceedings of the 2019 Conference on EMNLP-IJCNLP}, pages 1538--1548, Hong Kong, China. Association for Computational Linguistics.

\bibitem[Beltagy et~al., 2019]{scibert}
Beltagy, I., Lo, K., and Cohan, A. (2019).
\newblock Scibert: A pretrained language model for scientific text.
\newblock In {\em Conference on Empirical Methods in Natural Language Processing}.

\bibitem[Bodenreider, 2004]{umls}
Bodenreider, O. (2004).
\newblock The unified medical language system (umls): integrating biomedical terminology.
\newblock {\em Nucleic acids research}, 32.

\bibitem[Budzianowski et~al., 2018]{MultiWOZ}
Budzianowski, P., Wen, T.-H., Tseng, B.-H., Casanueva, I., Ultes, S., Ramadan, O., and Ga{\v{s}}i{\'c}, M. (2018).
\newblock {M}ulti{WOZ} - a large-scale multi-domain {W}izard-of-{O}z dataset for task-oriented dialogue modelling.
\newblock In {\em Proceedings of the 2018 Conference on Empirical Methods in Natural Language Processing}, pages 5016--5026, Brussels, Belgium. Association for Computational Linguistics.

\bibitem[Cai and Knight, 2013]{smatch}
Cai, S. and Knight, K. (2013).
\newblock {S}match: an evaluation metric for semantic feature structures.
\newblock In {\em Proceedings of the 51st Annual Meeting of the Association for Computational Linguistics (Volume 2: Short Papers)}, pages 748--752, Sofia, Bulgaria. ACL.

\bibitem[Chronopoulou et~al., 2022]{chronopoulou22}
Chronopoulou, A., Peters, M., and Dodge, J. (2022).
\newblock Efficient hierarchical domain adaptation for pretrained language models.
\newblock In {\em Proceedings of the 2022 Conference of the North American Chapter of the Association for Computational Linguistics: Human Language Technologies}, pages 1336--1351, Seattle, United States. Association for Computational Linguistics.

\bibitem[Chronopoulou et~al., 2023]{chronopoulou23}
Chronopoulou, A., Peters, M., Fraser, A., and Dodge, J. (2023).
\newblock {A}dapter{S}oup: Weight averaging to improve generalization of pretrained language models.
\newblock In {\em Findings of the Association for Computational Linguistics: EACL 2023}, pages 2054--2063, Dubrovnik, Croatia. Association for Computational Linguistics.

\bibitem[Cimini et~al., 2014]{cimini14}
Cimini, G., Gabrielli, A., and Labini, F. (2014).
\newblock The scientific competitiveness of nations.
\newblock {\em PloS one}, 9.

\bibitem[Colon-Hernandez et~al., 2021]{ColonHernandez21}
Colon-Hernandez, P., Havasi, C., Alonso, J.~B., Huggins, M., and Breazeal, C. (2021).
\newblock Combining pre-trained language models and structured knowledge.
\newblock {\em ArXiv}, abs/2101.12294.

\bibitem[Dettmers et~al., 2023]{dettmers2023qlora}
Dettmers, T., Pagnoni, A., Holtzman, A., and Zettlemoyer, L. (2023).
\newblock Qlora: Efficient finetuning of quantized llms.
\newblock {\em arXiv preprint arXiv:2305.14314}.

\bibitem[Diao et~al., 2023]{diao23}
Diao, S., Xu, T., Xu, R., Wang, J., and Zhang, T. (2023).
\newblock Mixture-of-domain-adapters: Decoupling and injecting domain knowledge to pre-trained language models{'} memories.
\newblock In {\em Proceedings of the 61st Annual Meeting of the Association for Computational Linguistics}, pages 5113--5129, Toronto, Canada. Association for Computational Linguistics.

\bibitem[Dinan et~al., 2018]{WoW}
Dinan, E., Roller, S., Shuster, K., Fan, A., Auli, M., and Weston, J. (2018).
\newblock Wizard of wikipedia: Knowledge-powered conversational agents.
\newblock {\em ArXiv}, abs/1811.01241.

\bibitem[Dogan et~al., 2014]{NCBI}
Dogan, R.~I., Leaman, R., and Lu, Z. (2014).
\newblock Ncbi disease corpus: A resource for disease name recognition and concept normalization.
\newblock {\em Journal of biomedical informatics}, 47:1--10.

\bibitem[Elsahar, 2017]{TREx}
Elsahar, H. (2017).
\newblock {T-Rex : A Large Scale Alignment of Natural Language with Knowledge Base Triples [NIF SAMPLE]}.

\bibitem[Emelin et~al., 2022]{emelin22}
Emelin, D., Bonadiman, D., Alqahtani, S., Zhang, Y., and Mansour, S. (2022).
\newblock Injecting domain knowledge in language models for task-oriented dialogue systems.
\newblock In {\em Proceedings of the 2022 Conference on Empirical Methods in Natural Language Processing}, pages 11962--11974. Association for Computational Linguistics.

\bibitem[Fichtl, 2024]{mythesis}
Fichtl, A. (2024).
\newblock Evaluating adapter-based knowledge-enhanced language models in the biomedical domain.
\newblock Master's thesis, Technical University of Munich, Munich, Germany.

\bibitem[Gu et~al., 2020]{blurb}
Gu, Y., Tinn, R., Cheng, H., Lucas, M.~R., Usuyama, N., Liu, X., Naumann, T., Gao, J., and Poon, H. (2020).
\newblock Domain-specific language model pretraining for biomedical natural language processing.
\newblock {\em ACM Transactions on Computing for Healthcare (HEALTH)}, 3:1 -- 23.

\bibitem[Guo and Guo, 2022]{guo22}
Guo, Q. and Guo, Y. (2022).
\newblock Lexicon enhanced chinese named entity recognition with pointer network.
\newblock {\em Neural Computing and Applications}.

\bibitem[Han et~al., 2021]{han2021}
Han, W., Pang, B., and Wu, Y.~N. (2021).
\newblock Robust transfer learning with pretrained language models through adapters.
\newblock {\em ArXiv}, abs/2108.02340.

\bibitem[He et~al., 2021a]{paralleladapter}
He, J., Zhou, C., Ma, X., Berg-Kirkpatrick, T., and Neubig, G. (2021a).
\newblock Towards a unified view of parameter-efficient transfer learning.
\newblock {\em ArXiv}, abs/2110.04366.

\bibitem[He et~al., 2021b]{he21}
He, R., Liu, L., Ye, H., Tan, Q., Ding, B., Cheng, L., Low, J.-W., Bing, L., and Si, L. (2021b).
\newblock On the effectiveness of adapter-based tuning for pretrained language model adaptation.

\bibitem[He et~al., 2020]{he20}
He, Y., Zhu, Z., Zhang, Y., Chen, Q., and Caverlee, J. (2020).
\newblock {I}nfusing {D}isease {K}nowledge into {BERT} for {H}ealth {Q}uestion {A}nswering, {M}edical {I}nference and {D}isease {N}ame {R}ecognition.
\newblock In {\em Proceedings of the 2020 Conference on Empirical Methods in Natural Language Processing (EMNLP)}, pages 4604--4614, Online. Association for Computational Linguistics.

\bibitem[Hogan et~al., 2020]{Hogan20}
Hogan, A., Blomqvist, E., Cochez, M., d’Amato, C., de~Melo, G., Guti{\'e}rrez, C., Kirrane, S., Gayo, J. E.~L., Navigli, R., Neumaier, S., Ngomo, A.-C.~N., Polleres, A., Rashid, S.~M., Rula, A., Schmelzeisen, L., Sequeda, J., Staab, S., and Zimmermann, A. (2020).
\newblock Knowledge graphs.
\newblock {\em ACM Computing Surveys (CSUR)}, 54:1 -- 37.

\bibitem[Houlsby et~al., 2019]{houlsbyadapter}
Houlsby, N., Giurgiu, A., Jastrzebski, S., Morrone, B., de~Laroussilhe, Q., Gesmundo, A., Attariyan, M., and Gelly, S. (2019).
\newblock Parameter-efficient transfer learning for nlp.
\newblock In {\em International Conference on Machine Learning}.

\bibitem[Hu et~al., 2022]{hu2022lora}
Hu, E.~J., Shen, Y., Wallis, P., Allen-Zhu, Z., Li, Y., Wang, S., Wang, L., and Chen, W. (2022).
\newblock Lo{RA}: Low-rank adaptation of large language models.
\newblock In {\em International Conference on Learning Representations}.

\bibitem[Hu et~al., 2023]{hu2023}
Hu, L., Liu, Z., Zhao, Z., Hou, L., Nie, L., and Li, J. (2023).
\newblock A survey of knowledge enhanced pre-trained language models.

\bibitem[{Huang} et~al., 2023]{huang_2023}
{Huang}, L., {Yu}, W., {Ma}, W., {Zhong}, W., {Feng}, Z., {Wang}, H., {Chen}, Q., {Peng}, W., {Feng}, X., {Qin}, B., and {Liu}, T. (2023).
\newblock {A Survey on Hallucination in Large Language Models: Principles, Taxonomy, Challenges, and Open Questions}.
\newblock {\em arXiv e-prints}, page arXiv:2311.05232.

\bibitem[Hung et~al., 2023]{hung23}
Hung, C.-C., Lange, L., and Str{\"o}tgen, J. (2023).
\newblock {TADA}: Efficient task-agnostic domain adaptation for transformers.
\newblock In {\em Findings of the Association for Computational Linguistics: ACL 2023}, pages 487--503, Toronto, Canada. Association for Computational Linguistics.

\bibitem[Hung et~al., 2022]{hung22}
Hung, C.-C., Lauscher, A., Ponzetto, S., and Glava{\v{s}}, G. (2022).
\newblock {DS}-{TOD}: Efficient domain specialization for task-oriented dialog.
\newblock In {\em Findings of the Association for Computational Linguistics: ACL 2022}, pages 891--904. Association for Computational Linguistics.

\bibitem[Ji et~al., 2020]{ji21}
Ji, S., Pan, S., Cambria, E., Marttinen, P., and Yu, P.~S. (2020).
\newblock A survey on knowledge graphs: Representation, acquisition, and applications.
\newblock {\em IEEE Transactions on Neural Networks and Learning Systems}, 33:494--514.

\bibitem[Jin et~al., 2019]{pubmedqa}
Jin, Q., Dhingra, B., Liu, Z., Cohen, W., and Lu, X. (2019).
\newblock {P}ub{M}ed{QA}: A dataset for biomedical research question answering.
\newblock In {\em Proceedings of the 2019 Conference on EMNLP-IJCNLP}, pages 2567--2577, Hong Kong, China. Association for Computational Linguistics.

\bibitem[K{\ae}r~J{\o}rgensen et~al., 2021]{kaer21}
K{\ae}r~J{\o}rgensen, R., Hartmann, M., Dai, X., and Elliott, D. (2021).
\newblock m{DAPT}: Multilingual domain adaptive pretraining in a single model.
\newblock In {\em Findings of the Association for Computational Linguistics: EMNLP 2021}, pages 3404--3418. Association for Computational Linguistics.

\bibitem[Khadir et~al., 2021]{KHADIR2021100339}
Khadir, A.~C., Aliane, H., and Guessoum, A. (2021).
\newblock Ontology learning: Grand tour and challenges.
\newblock {\em Computer Science Review}, 39:100339.

\bibitem[Kitchenham et~al., 2009]{kitchenham09}
Kitchenham, B., {Pearl Brereton}, O., Budgen, D., Turner, M., Bailey, J., and Linkman, S. (2009).
\newblock Systematic literature reviews in software engineering – a systematic literature review.
\newblock {\em Information and Software Technology}, 51(1):7--15.
\newblock Special Section - Most Cited Articles in 2002 and Regular Research Papers.

\bibitem[Lai et~al., 2023]{keblm}
Lai, T.~M., Zhai, C., and Ji, H. (2023).
\newblock Keblm: Knowledge-enhanced biomedical language models.
\newblock {\em Journal of Biomedical Informatics}, 143:104392.

\bibitem[Lauscher et~al., 2020]{lauscher20}
Lauscher, A., Majewska, O., Ribeiro, L. F.~R., Gurevych, I., Rozanov, N., and Glava{\v{s}}, G. (2020).
\newblock Common sense or world knowledge? investigating adapter-based knowledge injection into pretrained transformers.
\newblock In {\em Proceedings of Deep Learning Inside Out (DeeLIO)}, pages 43--49. Association for Computational Linguistics.

\bibitem[Lee et~al., 2019]{biobert}
Lee, J., Yoon, W., Kim, S., Kim, D., Kim, S., So, C.~H., and Kang, J. (2019).
\newblock {BioBERT: a pre-trained biomedical language representation model for biomedical text mining}.
\newblock {\em Bioinformatics}, 36(4):1234--1240.

\bibitem[Li et~al., 2023]{li23}
Li, B., Hwang, D., Huo, Z., Bai, J., Prakash, G., Sainath, T.~N., Chai~Sim, K., Zhang, Y., Han, W., Strohman, T., and Beaufays, F. (2023).
\newblock Efficient domain adaptation for speech foundation models.
\newblock In {\em ICASSP 2023 - 2023 IEEE International Conference on Acoustics, Speech and Signal Processing (ICASSP)}, pages 1--5.

\bibitem[Liu et~al., 2023]{liu23}
Liu, C., Zhang, S., Li, C., and Zhao, H. (2023).
\newblock Cpk-adapter: Infusing medical knowledge into k-adapter with continuous prompt.
\newblock In {\em 2023 8th International Conference on Intelligent Computing and Signal Processing (ICSP)}, pages 1017--1023, Los Alamitos, CA, USA. IEEE Computer Society.

\bibitem[Lu et~al., 2023]{lu23}
Lu, G., Yu, H., Yan, Z., and Xue, Y. (2023).
\newblock Commonsense knowledge graph-based adapter for aspect-level sentiment classification.
\newblock {\em Neurocomputing}, 534:67--76.

\bibitem[Lu et~al., 2021]{lu21}
Lu, Q., Dou, D., and Nguyen, T.~H. (2021).
\newblock Parameter-efficient domain knowledge integration from multiple sources for biomedical pre-trained language models.
\newblock In {\em Findings of the Association for Computational Linguistics: EMNLP 2021}, pages 3855--3865. Association for Computational Linguistics.

\bibitem[Majewska et~al., 2021]{majewska21}
Majewska, O., Vuli{\'c}, I., Glava{\v{s}}, G., Ponti, E.~M., and Korhonen, A. (2021).
\newblock Verb knowledge injection for multilingual event processing.
\newblock In {\em Proceedings of the 59th Annual Meeting of the ACL-IJCNLP}, pages 6952--6969. ACL.

\bibitem[Meng et~al., 2021]{mop}
Meng, Z., Liu, F., Clark, T.~H., Shareghi, E., and Collier, N. (2021).
\newblock Mixture-of-partitions: Infusing large biomedical knowledge graphs into bert.
\newblock {\em ArXiv}, abs/2109.04810.

\bibitem[Moon et~al., 2021]{moon21}
Moon, H., Park, C., Eo, S., Seo, J., and Lim, H. (2021).
\newblock An empirical study on automatic post editing for neural machine translation.
\newblock {\em IEEE Access}, 9:123754--123763.

\bibitem[Nentidis et~al., 2020]{bioasq7b}
Nentidis, A., Bougiatiotis, K., Krithara, A., and Paliouras, G. (2020).
\newblock Results of the seventh edition of the bioasq challenge.
\newblock In {\em Machine Learning and Knowledge Discovery in Databases}, pages 553--568, Cham. Springer International Publishing.

\bibitem[Nguyen-The et~al., 2023]{nguyen23}
Nguyen-The, M., Lamghari, S., Bilodeau, G.-A., and Rockemann, J. (2023).
\newblock Leveraging sentiment analysis knowledge to solve emotion detection tasks.
\newblock In {\em Pattern Recognition, Computer Vision, and Image Processing. ICPR 2022 International Workshops and Challenges}, pages 405--416. Springer Nature Switzerland.

\bibitem[Petroni et~al., 2019]{petroni19}
Petroni, F., Rockt{\"a}schel, T., Lewis, P., Bakhtin, A., Wu, Y., Miller, A.~H., and Riedel, S. (2019).
\newblock Language models as knowledge bases?
\newblock {\em ArXiv}, abs/1909.01066.

\bibitem[Pfeiffer et~al., 2020a]{pfeifferadapter}
Pfeiffer, J., Kamath, A., R{\"u}ckl{\'e}, A., Cho, K., and Gurevych, I. (2020a).
\newblock Adapterfusion: Non-destructive task composition for transfer learning.
\newblock {\em ArXiv}, abs/2005.00247.

\bibitem[Pfeiffer et~al., 2020b]{adapterhub}
Pfeiffer, J., R{\"u}ckl{\'e}, A., Poth, C., Kamath, A., Vuli{\'c}, I., Ruder, S., Cho, K., and Gurevych, I. (2020b).
\newblock Adapterhub: A framework for adapting transformers.
\newblock In {\em Proceedings of the 2020 Conference on Empirical Methods in Natural Language Processing: System Demonstrations}, pages 46--54.

\bibitem[Pfeiffer et~al., 2024]{pfeiffer2024modulardeeplearning}
Pfeiffer, J., Ruder, S., Vulić, I., and Ponti, E.~M. (2024).
\newblock Modular deep learning.
\newblock {\em Transcations of Machine Learning Research}.

\bibitem[Qian et~al., 2022]{qian22}
Qian, Y., Gong, X., and Huang, H. (2022).
\newblock Layer-wise fast adaptation for end-to-end multi-accent speech recognition.
\newblock {\em IEEE/ACM Transactions on Audio, Speech, and Language Processing}, 30:2842--2853.

\bibitem[Rebuffi et~al., 2017]{rebuffi17}
Rebuffi, S.-A., Bilen, H., and Vedaldi, A. (2017).
\newblock Learning multiple visual domains with residual adapters.
\newblock In {\em Advances in Neural Information Processing Systems 30: Annual Conference on Neural Information Processing Systems 2017}, pages 506--516.

\bibitem[Romanov and Shivade, 2018]{mednli}
Romanov, A. and Shivade, C. (2018).
\newblock Lessons from natural language inference in the clinical domain.
\newblock In {\em Proceedings of the 2018 Conference on Empirical Methods in Natural Language Processing}, pages 1586--1596, Brussels, Belgium. ACL.

\bibitem[{Rosset} et~al., 2020]{rosset_2020}
{Rosset}, C., {Xiong}, C., {Phan}, M., {Song}, X., {Bennett}, P., and {Tiwary}, S. (2020).
\newblock {Knowledge-Aware Language Model Pretraining}.
\newblock {\em arXiv e-prints}, page arXiv:2007.00655.

\bibitem[Schabus et~al., 2017]{omp}
Schabus, D., Skowron, M., and Trapp, M. (2017).
\newblock One million posts: A data set of german online discussions.
\newblock In {\em Proceedings of the 40th International ACM SIGIR Conference on Research and Development in Information Retrieval}, page 1241–1244, New York, NY, USA. Association for Computing Machinery.

\bibitem[Schneider et~al., 2022]{schneider22}
Schneider, P., Schopf, T., Vladika, J., Galkin, M., Simperl, E., and Matthes, F. (2022).
\newblock A decade of knowledge graphs in natural language processing: A survey.
\newblock In {\em Proceedings of the 2nd AACL-IJCNLP}, pages 601--614. Association for Computational Linguistics.

\bibitem[Schuler, 2006]{VerbNet}
Schuler, K.~K. (2006).
\newblock {\em VerbNet: A Broad-Coverage, Comprehensive Verb Lexicon}.
\newblock PhD thesis, University of Pennsylvania.

\bibitem[Shi et~al., 2021]{AESRC2020}
Shi, X., Yu, F., Lu, Y., Liang, Y., Feng, Q., Wang, D., Qian, Y., and Xie, L. (2021).
\newblock The accented english speech recognition challenge 2020: Open datasets, tracks, baselines, results and methods.
\newblock {\em ICASSP 2021 - IEEE International Conference on Acoustics, Speech and Signal Processing}, pages 6918--6922.

\bibitem[Speer et~al., 2017]{speer17}
Speer, R., Chin, J., and Havasi, C. (2017).
\newblock Conceptnet 5.5: An open multilingual graph of general knowledge.
\newblock In {\em Proceedings of the Thirty-First AAAI Conference on Artificial Intelligence}. AAAI Press.

\bibitem[Stickland and Murray, 2019]{stickland19}
Stickland, A.~C. and Murray, I. (2019).
\newblock {BERT} and {PAL}s: Projected attention layers for efficient adaptation in multi-task learning.
\newblock In {\em Proceedings of the 36th International Conference on Machine Learning}, pages 5986--5995. PMLR.

\bibitem[Tian et~al., 2024]{tian-etal-2024-kg}
Tian, S., Luo, Y., Xu, T., Yuan, C., Jiang, H., Wei, C., and Wang, X. (2024).
\newblock {KG}-adapter: Enabling knowledge graph integration in large language models through parameter-efficient fine-tuning.
\newblock In {\em Findings of the Association for Computational Linguistics ACL 2024}, pages 3813--3828, Bangkok, Thailand and virtual meeting. Association for Computational Linguistics.

\bibitem[Tiwari et~al., 2022]{VisMDD}
Tiwari, A., Manthena, M., Saha, S., Bhattacharyya, P., Dhar, M., and Tiwari, S. (2022).
\newblock Dr. can see: Towards a multi-modal disease diagnosis virtual assistant.
\newblock In {\em Proceedings of the 31st ACM International Conference on Information \& Knowledge Management}, page 1935–1944, New York, NY, USA. Association for Computing Machinery.

\bibitem[Tiwari et~al., 2023]{tiwari23}
Tiwari, A., Saha, A., Saha, S., Bhattacharyya, P., and Dhar, M. (2023).
\newblock Experience and evidence are the eyes of an excellent summarizer! towards knowledge infused multi-modal clinical conversation summarization.
\newblock In {\em Proceedings of the 32nd ACM International Conference on Information and Knowledge Management}, page 2452–2461. Association for Computing Machinery.

\bibitem[Vaswani et~al., 2017]{vaswani2017}
Vaswani, A., Shazeer, N.~M., Parmar, N., Uszkoreit, J., Jones, L., Gomez, A.~N., Kaiser, L., and Polosukhin, I. (2017).
\newblock Attention is all you need.
\newblock In {\em NIPS}.

\bibitem[Vasylenko et~al., 2023]{vasylenko23}
Vasylenko, P., Huguet~Cabot, P.~L., Mart{\'\i}nez~Lorenzo, A.~C., and Navigli, R. (2023).
\newblock Incorporating graph information in transformer-based {AMR} parsing.
\newblock In {\em Findings of the Association for Computational Linguistics: ACL 2023}, pages 1995--2011, Toronto, Canada. Association for Computational Linguistics.

\bibitem[Vladika. et~al., 2024]{icaart24}
Vladika., J., Fichtl., A., and Matthes., F. (2024).
\newblock Diversifying knowledge enhancement of biomedical language models using adapter modules and knowledge graphs.
\newblock In {\em Proceedings of the 16th International Conference on Agents and Artificial Intelligence - Volume 2: ICAART}, pages 376--387. INSTICC, SciTePress.

\bibitem[Wang et~al., 2019]{glue}
Wang, A., Singh, A., Michael, J., Hill, F., Levy, O., and Bowman, S.~R. (2019).
\newblock Glue: A multi-task benchmark and analysis platform for natural language understanding.

\bibitem[Wang et~al., 2020]{kadapter}
Wang, R., Tang, D., Duan, N., Wei, Z., Huang, X., Ji, J., Cao, G., Jiang, D., and Zhou, M. (2020).
\newblock K-adapter: Infusing knowledge into pre-trained models with adapters.

\bibitem[Wei et~al., 2021]{Wei21}
Wei, X., Wang, S., Zhang, D., Bhatia, P., and Arnold, A.~O. (2021).
\newblock Knowledge enhanced pretrained language models: A compreshensive survey.
\newblock {\em ArXiv}, abs/2110.08455.

\bibitem[Wenzek et~al., 2020]{CCNet}
Wenzek, G., Lachaux, M.-A., Conneau, A., Chaudhary, V., Guzm{\'a}n, F., Joulin, A., and Grave, E. (2020).
\newblock {CCN}et: Extracting high quality monolingual datasets from web crawl data.
\newblock In {\em Proceedings of the Twelfth Language Resources and Evaluation Conference}, pages 4003--4012.

\bibitem[Wold, 2022]{wold22}
Wold, S. (2022).
\newblock The effectiveness of masked language modeling and adapters for factual knowledge injection.
\newblock In {\em Proceedings of TextGraphs-16}, pages 54--59, Gyeongju, Republic of Korea. Association for Computational Linguistics.

\bibitem[Xie et~al., 2022]{kebiosum}
Xie, Q., Bishop, J.~A., Tiwari, P., and Ananiadou, S. (2022).
\newblock Pre-trained language models with domain knowledge for biomedical extractive summarization.
\newblock {\em Knowledge-Based Systems}, 252:109460.

\bibitem[Xu et~al., 2022]{xu22}
Xu, Y., Ishii, E., Cahyawijaya, S., Liu, Z., Winata, G.~I., Madotto, A., Su, D., and Fung, P. (2022).
\newblock Retrieval-free knowledge-grounded dialogue response generation with adapters.
\newblock In {\em Proceedings of the Second DialDoc Workshop on Document-grounded Dialogue and Conversational Question Answering}, pages 93--107. Association for Computational Linguistics.

\bibitem[Yu and Yang, 2023]{yu23}
Yu, S. and Yang, Y. (2023).
\newblock A new feature fusion method based on pre-training model for sequence labeling.
\newblock In {\em 2023 6th International Conference on Data Storage and Data Engineering (DSDE)}, pages 26--31.

\bibitem[Zou et~al., 2022]{zou22}
Zou, D., Zhang, X., Song, X., Yu, Y., Yang, Y., and Xi, K. (2022).
\newblock Multiway bidirectional attention and external knowledge for multiple-choice reading comprehension.
\newblock In {\em 2022 IEEE International Conference on Systems, Man, and Cybernetics (SMC)}, pages 694--699.

\end{thebibliography}
\bibliographystyle{apalike}


\appendix
\setitemize{noitemsep,topsep=0.5pt,parsep=0pt,partopsep=0pt}
\section{Supplementary Survey Data}
\label{sec:appendix}

\subsection{Methodology}
\label{sec:appendix_methodology}

\textbf{Articles on the following topics were excluded:}

\begin{itemize}
\setlength\itemsep{0em}
    \item Articles published before February 2, 2019
    \item Duplicate versions of the same article (when multiple versions of an article were found in different journals, only the most recent version was included)
    \item Articles where Adapters were used for NLP, but for use-cases other than knowledge-enhancement (such as few-shot learning or model debiasing)
    \item Articles written in a language other than English
\end{itemize}

\textbf{Extracted data:}

\begin{itemize}
\setlength\itemsep{0em}
    \item Publication source, date, and full reference
    \item Main topic area
    \item Facts of interest such as adapter architecture, domain, and downstream tasks within the papers
    \item A short summary of the study, including the main research questions and the answers
\end{itemize}

\subsection{Acronyms}
\label{sec:appendix_acronyms}

\begin{itemize}
\setlength\itemsep{0em}
    \item{AESRC2020: Accented English Speech Recognition Challenge 2020 \citep{AESRC2020}}
    \item{BLURB: Biomedical Language Understanding and Reasoning Benchmark \citep{blurb}}
    \item{CCNet: Common Crawl Net \citep{CCNet}}
    \item{EE: Event Extraction}
    \item{EL: Entity Linking}
    \item{ES: Extractive Summarization}
    \item{ET: Entity Typing}
    \item{GLUE: General Language Understanding Evaluation \citep{glue}}
    \item{IE: Information Extraction}
    \item{KGD: Knowledge-grounded Dialogue}
    \item{LAMA: Concept-Net Split of LAMA Probe \citep{petroni19}}
    \item{LM: Language Modeling}
    \item{MT: Machine Translation}
    \item{MultiWOZ: Multi-Domain Wizard-of-Oz dataset \citep{MultiWOZ}}
    \item{NER: Named Entity Recognition}
    \item{NLI: Natural Language Inference}
    \item{OOD: Out-of-domain Detection}
    \item{QA: Question Answering}
    \item{RC: Reading Comprehension}
    \item{RE: Relation Extraction}
    \item{RCL: Relation Classification}
    \item{SA: Sentiment Analysis}
    \item{SC: Sentiment Classification}
    \item{SF: Speech Foundation}
    \item{SL: Sequence Labelling}
    \item{SMATCH: Semantic Match Score \citep{smatch}}
    \item{SR: Speech Recognition}
    \item{STC: Sentence Classification}
    \item{TC: Text Classification}
    \item{TOD: Task-Oriented dialogue}
    \item{T-REx: A Large Scale Alignment of Natural Language with Knowledge Base Triples \citep{TREx}}
    \item{VerbNet: A Broad-Coverage, Comprehensive Verb Lexicon \citep{VerbNet}}
    \item{Vis-MDD: Visual Medical Disease Diagnosis \citep{VisMDD}}
    \item{WMT20: Workshop on Machine Translation 2020} 
    \item{WoW: Wizard-of-Wikipedia \citep{WoW}}
\end{itemize}

\newpage 
\newpage 


\end{document}